\documentclass{article}
\PassOptionsToPackage{numbers,sort&compress}{natbib}
\usepackage[dblblindworkshop,final]{neurips_2026}
\workshoptitle{AI for Stochastic Dynamics}

\usepackage[utf8]{inputenc}
\usepackage[T1]{fontenc}
\usepackage{hyperref}
\usepackage{url}
\usepackage{booktabs}
\usepackage{graphicx}
\usepackage{amsfonts}
\usepackage{amsmath}
\usepackage{amssymb}
\usepackage{mathtools}
\usepackage{microtype}
\usepackage{xcolor}
\usepackage{tabularx}
\usepackage{array}
\usepackage{enumitem}
\usepackage{amsthm}

\newtheorem{proposition}{Proposition}

\newcommand{\R}{\mathbb{R}}
\newcommand{\E}{\mathbb{E}}

\newcommand{\cL}{\mathcal{L}}
\newcommand{\dd}{\mathrm{d}}

\title{Probability flow ODEs in \\
score-based  and reflected diffusion models}
\author{Rama CONT\\
Mathematical Institute, University of Oxford}

\begin{document}

\maketitle

\begin{abstract}
Probability-flow ordinary differential equations (PF-ODEs) are widely used as deterministic samplers for score-based diffusion models. Their usual justification is that the Fokker--Planck equation of a diffusion can be rewritten as a continuity equation driven by the score function of the forward diffusion. This identity does not, however, guarantee that the resulting velocity field generates a well-posed  flow. We provide theoretical insights into the design of such deterministic samplers for generative models based on diffusions and reflected diffusions. We identify  sufficient conditions for a regular Lagrangian PF-ODE flow to exist; reverse sampling and invertibility require two-sided divergence control. For learned scores, sampler stability is controlled by an unweighted velocity error, exposing a mismatch with density-weighted score matching and motivating architectural control of Jacobians, divergence, growth, and compression. Under the manifold hypothesis, positive-time regularization justifies an early-stopped PF-ODE while constants deteriorate near the data endpoint; an explicit sphere example shows that the exact deterministic flow becomes singular as the noise level vanishes even though the diffusion marginals remain well defined. 
  
These theoretical insights translate into concrete design principles for stable, invertible, and constraint-preserving diffusion samplers. We illustrate the practical relevance of these design principles using controlled numerical experiments.
\end{abstract}
\section{Probability-flow ODEs in diffusion models}
Score-based diffusion models \citep{song2021sde} generate data by reversing a  noising process and may be sampled either with a reverse-time SDE \citep{anderson1982,cattiaux1988time} or with the deterministic probability-flow ODE (PF-ODE) introduced in \citet{song2021ml}. The PF-ODE is attractive because it removes stochastic simulation, supports deterministic encoding/decoding, and   permits   likelihood calculations \citep{song2021sde,song2021ml}. In constrained generative modeling, reflected diffusion models \citep{lou2023reflected,fishman2023constrained} keep the forward process inside a prescribed domain and can likewise be paired with a deterministic PF-ODE.

The formal derivation of the PF-ODE in these references is based on the observation that, for the diffusion
\begin{equation}
  \dd X_t=f(X_t,t)\,\dd t+\sigma(t)\,\dd W_t,
  \label{eq:sde}
\end{equation}
the marginal density $p_t$ satisfies the Fokker--Planck equation
\begin{equation}
  \partial_t p_t+\nabla\!\cdot(f p_t)-\tfrac12\sigma(t)^2\Delta p_t=0.
  \label{eq:fp}
\end{equation}
Introducing the score function $s_t=\nabla\log p_t$, \eqref{eq:fp} can be written as
\begin{equation}
 \partial_t p_t+\nabla\cdot(p_t v)=0\quad{\rm where}\quad v(x,t)=f(x,t)-\tfrac12\sigma(t)^2 s_t(x)
  \label{eq:velocity}
\end{equation}
which is the {\it continuity equation} for the ordinary differential equation (ODE)  $\dot Z_t=v(Z_t,t)$ and concludes that $Z_t$ has density $p_t$.
It is then tempting to conclude that the ODE $\dot Z_t=v(Z_t,t)$ automatically transports
$p_0$ to $p_t$. This conclusion is not automatic \citep{cont2026mimicking}: the PDE  \eqref{eq:fp} determines an
\emph{Eulerian} density evolution, whereas a deterministic sampler requires a
\emph{Lagrangian} flow of individual samples. A unique and smooth density flow can exist
even when the associated PF-ODE fails to generate a regular Lagrangian flow from the data
time \citep{cont2026mimicking}.
This distinction is especially relevant under the manifold hypothesis \citep{debortoli2022manifold} i.e. when data law is  supported on a lower-dimensional manifold.

\paragraph{Contributions.} We use theoretical insights from
\cite{cont2026mimicking,cont2026reflected} to derive implications for the design and training of score-based and reflected diffusion models, and exhibit   explicit  examples
and sampler diagnostics.  First, when data is supported on a lower dimensional manifold, positive-time diffusion regularizes the singular data law but the
PF-ODE may become singular as $t\downarrow0$. In Section \ref{sec:manifold} we give an explicit  example of this type, which leads to
collapse of the normal directions at the data endpoint. This complements convergence results
for diffusion models under the manifold hypothesis
\citep{debortoli2022manifold,azangulov2025manifold,potaptchik2025linear}.
Second, for learned scores, deterministic sampler stability depends not only on score-matching
error but also on regularity, divergence, growth, and compression of the induced velocity
field. Third, for reflected models, the PF-ODE contains no explicit reflection term: 
confinement follows from tangency of the  probability-flow
velocity, rather than from a zero-normal-score condition.

These observations lead to a common design principle: the PF-ODE should be treated as a
learned dynamical system with its own well-posedness and conditioning requirements, rather
than as an automatic consequence of score estimation. In particular, early stopping,
velocity-field regularization, divergence control, low-density validation, and appropriate
boundary constraints should be viewed as structural components of deterministic diffusion
sampling.
We illustrate these principles with 
numerical experiments (Section~\ref{sec:design} and Appendix \ref{sec:numerics}), with a focus on endpoint conditioning, the
DSM--velocity-error gap, and  Jacobian regularization.

\section{From diffusion marginals to a deterministic sampler}
\label{sec:pfode}

\subsection{Marginal matching requires a Lagrangian flow}

The identity between \eqref{eq:fp} and the continuity equation establishes that $p_t$ is an Eulerian solution for the velocity \eqref{eq:velocity}. To obtain a sampler one needs a flow map $Z(t,\cdot)$ whose trajectories solve $\dot Z=v(Z,t)$ which transports $p_0$ to $p_t$.
\begin{equation}
  Z(t,\cdot)_{\#}(p_0\,\dd x)=p_t\,\dd x.
  \label{eq:push}
\end{equation}
For non-smooth velocity fields the relevant object is a regular Lagrangian flow (RLF), which permits Sobolev or bounded-variation (BV) spatial regularity while imposing a quantitative compressibility bound. Under local BV regularity of the drift and score, linear growth, and a one-sided bound on the negative divergence of $v$, such a unique RLF exists and satisfies \eqref{eq:push} \citep{cont2026mimicking,diperna1989,ambrosio2004}. This is substantially weaker than assuming a globally Lipschitz score.

The distinction matters for model validation. A well-posed SDE and a unique Fokker--Planck density do not certify the PF-ODE. The diffusion benefits from stochastic regularization, whereas the deterministic dynamics directly inherit score singularities. For a Brownian forward process initialized by the uniform law on an interval, for example, $p_t$ is smooth and strictly positive for every $t>0$, yet the score outside the initial support behaves as $O(1/t)$ as $t\downarrow0$; the PF-ODE has no regular Lagrangian flow from time zero \citep{cont2026mimicking}. Thus ``the score is smooth for every positive time'' is weaker than the time-integrated regularity needed by the sampler.

\paragraph{Design implication: early stopping is structural.}
Under the manifold hypothesis linear VP/VE forward diffusions regularize the law at every $\delta>0$. On $[\delta,T]$ the score has explicit Hessian, divergence, and linear-growth bounds sufficient for PF-ODE well-posedness, with constants that deteriorate as $\delta\downarrow0$ \citep{cont2026mimicking}. The principled deterministic pipeline is therefore
\begin{equation}
  \mu_0 \xrightarrow{\text{forward diffusion}} p_\delta
  \xrightarrow{\text{PF-ODE on }[\delta,T]} p_T,
  \qquad
  p_T \xrightarrow{\text{reverse PF-ODE}} p_\delta,
  \label{eq:earlystop}
\end{equation}
and the final passage $p_\delta\to\mu_0$ requires a limiting argument or a separate terminal denoising step. Reducing $\delta$ improves endpoint fidelity but worsens conditioning. This gives an analytic interpretation of the empirical difficulty of the final denoising steps and suggests treating the truncation time as a model-design parameter, not merely an ODE-solver tolerance.

\subsection{Reverse sampling is stronger than forward marginal matching}

The RLF criterion for forward transport uses a one-sided compression bound, schematically
\begin{equation}
  [\nabla\cdot v]^-\in L^1_tL^\infty_x.
  \label{eq:negdiv}
\end{equation}
Reversing time sends $v$ to $-v$ and exchanges the positive and negative parts of the divergence. Consequently, a reverse deterministic sampler additionally needs
\begin{equation}
  [\nabla\cdot v]^+\in L^1_tL^\infty_x.
  \label{eq:posdiv}
\end{equation}
Under both bounds the flow is essentially invertible, the reverse RLF transports $p_T$ to $p_0$ (or to $p_\delta$ in the early-stopped setting), and encode--decode inversion and change-of-variables likelihoods become justified \citep{cont2026mimicking}.
This asymmetry matters for diffusion-model objectives. Forward marginal matching, reverse deterministic generation, and exact likelihood evaluation are not equivalent specifications. A model intended only for one-way generation may tolerate weaker divergence control than a model intended as an invertible continuous normalizing flow. Architectural and regularization choices should therefore be aligned with the required downstream operation.

\section{Singularity at the data endpoint under the manifold hypothesis}
\label{sec:manifold}

By \emph{data endpoint} we mean the zero-noise time $t=0$, not the boundary of a spatial
domain. The manifold hypothesis posits that high-dimensional data concentrate on, or near,
a set whose intrinsic dimension is much smaller than the ambient dimension. We focus here
on the idealized but important limiting case in which the data distribution is \emph{exactly}
supported on a lower-dimensional manifold and is therefore singular with respect to ambient
Lebesgue measure. The conclusions below need not hold in the same form for distributions
that merely concentrate near a manifold while retaining a positive ambient density.

Diffusion models can nevertheless be well behaved in this singular setting. Convergence to
targets supported on lower-dimensional manifolds or empirical measures has been established
without assuming an ambient target density \citep{debortoli2022manifold}; more recent results
show that score-learning and sampling rates can depend primarily on intrinsic rather than
ambient dimension \citep{azangulov2025manifold,potaptchik2025linear}. The question addressed
here is different: does such favorable distributional behavior imply that the deterministic
probability-flow ODE remains a regular and invertible ambient-space flow all the way to
$t=0$? The following example shows that it does not.

The construction is a higher-dimensional version of the time-zero vacuum mechanism in
\cite{cont2026mimicking}: diffusion immediately fills directions in which the initial law has
no ambient density, while the corresponding deterministic probability-flow velocity becomes
singular as this regularization is removed.

\begin{proposition}[Sphere data: singularity at the data endpoint]
\label{prop:sphere}
Let $\mu_0$ be the uniform distribution on
\[
  \mathcal M
  =
  S_R^m\times\{0\}
  \subset
  \R^{m+1}\times\R^{d-m-1}
  =
  \R^d,
  \qquad m<d,
\]
and consider the variance-exploding forward diffusion
\[
  X_t=X_0+W_t,
  \qquad X_0\sim\mu_0.
\]
For every $t>0$, $X_t$ has a smooth strictly positive density $p_t$.

If $d\ge m+2$, write $x=(u,z)\in\R^{m+1}\times \R^{d-m-1}$. Then the velocity
$v=-\frac12\nabla\log p_t$ satisfies  
\begin{equation}
  v_z(u,z,t)=\frac{z}{2t}.
  \label{eq:normalvelocity}
\end{equation}
Consequently, no absolutely continuous PF-ODE trajectory on $[0,T]$ can start from a point
with $z_0\neq0$. Since $\{z\neq0\}$ has full ambient Lebesgue measure, there is no regular
Lagrangian flow from $t=0$ with respect to ambient Lebesgue measure.

In reverse time, however, a trajectory initialized at $T>0$ satisfies
\begin{equation}
  z_t=\sqrt{\frac{t}{T}}\,z_T,
  \label{eq:normalcollapse}
\end{equation}
and hence converges continuously to the lower-dimensional subspace as $t\downarrow0$.
Thus the obstruction at the data endpoint concerns ambient-flow regularity, compression,
and invertibility; it does not imply nonexistence of individual reverse trajectories.
\end{proposition}

\paragraph{Derivation.}
Because the last $d-m-1$ coordinates of $X_0$ vanish and Brownian noise is isotropic,
\begin{equation}
  p_t(u,z)
  =
  q_t(u)(2\pi t)^{-(d-m-1)/2}
  \exp\!\left(-\frac{|z|^2}{2t}\right),
  \label{eq:spherefactorization}
\end{equation}
where $q_t$ is the Gaussian smoothing of the uniform law on
$S_R^m\subset\R^{m+1}$. Therefore
\[
  \nabla_z\log p_t(u,z)=-\frac{z}{t},
\]
which proves \eqref{eq:normalvelocity}. Every solution of
$\dot z=z/(2t)$ on $(0,T]$ has the form
$  z_t=C\sqrt{t}.$
Continuity at $t=0$ therefore forces $z_0=0$, so an ambient trajectory cannot be launched
at time zero from Lebesgue-a.e.\ point. Conversely, fixing $z_T$ yields
\eqref{eq:normalcollapse}. In the $c=d-m-1$ displayed normal directions, the reverse map
contracts volume by
\[
  \left(\frac{t}{T}\right)^{c/2},
\]
which tends to zero at the data endpoint.

The conclusion concerns an \emph{ambient} regular Lagrangian flow, whose reference measure
is $d$-dimensional Lebesgue measure. Since $\mu_0$ itself is concentrated on a Lebesgue-null
set, Proposition~\ref{prop:sphere} does not rule out a deterministic transport defined only
$\mu_0$-a.s.\ on $\mathcal M$. Rather, it shows that such a transport cannot remain a uniformly
regular, invertible ambient-space flow as the noise level vanishes.

\paragraph{Codimension one.}
The case $d=m+1$, where there are no additional flat normal coordinates, exhibits the same
mechanism in the radial normal direction. Writing $r=|x|$ and
$\nu=(m-1)/2$, the radial score of the Gaussian smoothing of the uniform law on $S_R^m$
can be written as
\begin{equation}
  \partial_r\log p_t(r)
  =
  -\frac{r}{t}
  +
  \frac{R}{t}
  \frac{I_{\nu+1}(Rr/t)}{I_\nu(Rr/t)},
  \label{eq:besselscore}
\end{equation}
where $I_\nu$ denotes the modified Bessel function. Since
\[
  \frac{I_{\nu+1}(a)}{I_\nu(a)}\longrightarrow1,
  \qquad a\to\infty,
\]
we obtain, uniformly on compact subsets of
$\{x\in\R^{m+1}:0<|x|\neq R\}$,
\begin{equation}
  \partial_r\log p_t(r)
  =
  -\frac{r-R}{t}+O(1),
  \qquad t\downarrow0.
  \label{eq:radialasymptotic}
\end{equation}
Thus, at every fixed ambient point separated from the sphere and from the origin, the normal
component of the probability-flow velocity has a $1/t$ blow-up. If an absolutely continuous
trajectory were continuous at $t=0$ with $r_0\neq R$, then for sufficiently small $t$ its
distance from the sphere would remain bounded away from zero, making the time integral of
the leading velocity term divergent. Hence such a trajectory cannot start at that point.
This again excludes an ambient regular Lagrangian flow from the data endpoint.

Notice that this statement does not conflict with reverse-time collapse onto the sphere.
Along a reverse trajectory the normal distance itself decreases with $t$, so the velocity
experienced along that trajectory need not scale as $1/t$. The $1/t$ statement concerns
the velocity  away from the data manifold.

\paragraph{Relation to convergence results under the manifold hypothesis.}
\citet{debortoli2022manifold} showed that denoising diffusion models can converge to targets
supported on lower-dimensional manifolds or empirical measures and obtained quantitative
$W_1$ guarantees without assuming an ambient target density.
\citet{azangulov2025manifold} obtain score-learning and reverse-SDE sampling rates largely independent of ambient dimension,  while
\citet{potaptchik2025linear} showed that the number of reverse-SDE discretization steps required for KL convergence can scale linearly, up to logarithmic factors, with intrinsic dimension. These are distributional convergence results for stochastic diffusion samplers.

 Proposition~\ref{prop:sphere} addresses a different question: whether the associated deterministic probability flow remains a regular and invertible ambient-space flow as the noise level vanishes.  The sphere
example shows that, even with the \emph{exact} score, approaching a lower-dimensional target forces collapse of normal directions. Thus favorable intrinsic-dimensional sampling rates do not imply  Lagrangian regularity, bounded compression or invertibility of the PF-ODE at $t=0$.
\paragraph{Design implication.}
Under the manifold hypothesis, when the target distribution is idealized as being exactly
supported on a lower-dimensional manifold, the zero-noise limit should be viewed as a
\emph{singular collapse limit}, rather than as an ordinary well-conditioned ODE time.
If stable deterministic transport, invertible encoding/decoding, or change-of-variables
likelihood evaluation is required, a natural strategy is to stop the PF-ODE at some
$\delta>0$ and treat the  passage from $p_\delta$ to the data distribution separately,
for example through a terminal denoising or manifold-valued correction step.

If reverse trajectories are integrated closer to zero, one should expect normal Jacobian
factors and ambient volume elements to collapse, while the conditioning constants of the
deterministic flow deteriorate even for a perfect score. This is an intrinsic geometric
consequence of removing diffusion regularization, not merely an effect of score-estimation
error.
\section{Learned scores: training error versus sampler stability}
\label{sec:learned}

Let $s_\theta$ be a learned approximation of $s=\nabla\log p$ and
\begin{equation}
  v_\theta=f-\tfrac12\sigma^2s_\theta.
\end{equation}
On an early-stopped interval $[\delta,T]$, the RLF stability theory gives trajectory convergence under local Sobolev regularity, a uniform compression bound, and convergence of the velocity in an unweighted $L^1$ norm. Writing
\begin{equation}
 E_v:=\|v-v_\theta\|_{L^1([\delta,T]\times B_R)},
 \label{eq:Ev}
\end{equation}
the measure of initial points whose exact and learned trajectories differ by more than $\eta$ admits a logarithmic modulus in $E_v$ for Sobolev velocities \citep{cont2026mimicking,crippa2008}. In the globally Lipschitz regime, a standard Gr\"onwall argument yields a linear dependence on a uniform score error, but with an amplification factor that can grow rapidly as $\delta\downarrow0$.
The important   point is that $E_v$ is not the standard denoising score-matching objective
\begin{equation}
  E_\theta^2
  :=\int_\delta^T\!\int p_t(x)
       \left|s_\theta(x,t)-s_t(x)\right|^2\,\dd x\,\dd t.
  \label{eq:scoreloss}
\end{equation}
On a ball $B_R$, Cauchy--Schwarz gives the conversion
\begin{equation}
  E_v \le \frac12\|\sigma\|_\infty^2 E_\theta\sqrt{\kappa_{\delta,R}},
  \qquad
  \kappa_{\delta,R}:=\int_\delta^T\!\int_{B_R}\frac{1}{p_t(x)}\,\dd x\,\dd t.
  \label{eq:mismatch}
\end{equation}
Thus low-density regions receive little weight in \eqref{eq:scoreloss} but are amplified in the conversion to the velocity norm controlling deterministic trajectories. For standard linear VP/VE diffusions, $\kappa_{\delta,R}$ is finite for every $\delta>0$ and finite $R$, but can grow sharply as $\delta\downarrow0$ or $R$ increases \citep{cont2026mimicking}. This is a concrete reason why a small score loss need not imply a stable PF-ODE sampler.

\paragraph{Design implication: architecture should enforce regularity.}
The learned flow must control more than pointwise error. Sufficient conditions include local $W^{1,p}$ regularity ($p>1$), a one-sided divergence bound for $v_\theta$, and linear growth. These quantities are not controlled by density-weighted score matching alone. Practical mechanisms suggested by the theory include spectral normalization, Jacobian  penalties, explicit divergence regularization, and uniform control of network growth. For a sequence of increasingly accurate score models,  compression and growth constants must remain uniformly bounded; otherwise $E_\theta\to0$ does not imply convergence of the samplers.
This implication complements statistical score-estimation guarantees \citep{han2024score} and convergence analyses that assume smooth PF-ODE dynamics \citep{huang2025pfode} and
 also connects to schedule design \citep{chen2025schedule,mooney2025wellposed}. 
 A schedule optimized only for score-estimation difficulty or terminal SNR may be poorly conditioned for deterministic transport near the data endpoint.
\vspace{-10pt}
\section{Reflected diffusion models: deterministic confinement without reflection}
\label{sec:reflected}
Constrained diffusions \citep{lou2023reflected,fishman2023constrained} replace \eqref{eq:sde} by a normally reflected diffusion in a bounded domain:
\begin{equation}
  \dd X_t=f(X_t,t)\,\dd t+\sigma(t)\,\dd W_t-\nu(X_t)\,\dd L_t,
  \qquad X_t\in\overline\Omega\subset\R^d,
  \label{eq:reflectedsde}
\end{equation}
where $\nu$ is the outward unit normal and $L_t$ is boundary local time. Reflected diffusion models use such forward dynamics to prevent mass from leaving the feasible set \citep{lou2023reflected,fishman2023constrained}. Their marginal density satisfies a Fokker--Planck equation with co-normal no-flux condition
\begin{equation}
  J\cdot\nu=0,
  \qquad
  J:=fp-\tfrac12\sigma^2\nabla p.
  \label{eq:noflux}
\end{equation}
Define again $v=J/p=f-\frac12\sigma^2\nabla\log p$. If $p>0$  is differentiable up to the boundary \eqref{eq:noflux} implies
\begin{equation}
  v\cdot\nu=0\quad\text{on }\partial\Omega.
  \label{eq:tangent}
\end{equation}
The  local-time term has disappeared: its effect on the marginal flow is encoded by the boundary behavior of the score. Under positive-time regularization and   drift--score regularity, the  ODE
\begin{equation}
  \dot Z_t=f(Z_t,t)-\tfrac12\sigma(t)^2\nabla\log p_t(Z_t)
  \label{eq:refpfode}
\end{equation}
generates a confined regular Lagrangian flow and transports the same marginals as the reflected diffusion \citep{cont2026reflected}. In particular, for the reflected VE   ($f=0$) and reflected VP models ($f(x,t)=-\beta(t)x$, $\sigma^2=2\beta$), this conclusion holds on every $[\delta,T]$ for arbitrary initial laws in a   smooth bounded domain.
\vspace{-20pt}
\subsection{The boundary condition acts on the full velocity}
Equation \eqref{eq:tangent} is the correct design target. It does \emph{not} generally mean that the normal score is zero. Dividing the no-flux condition by $p>0$ gives
\begin{equation}
  \partial_\nu\log p = \frac{2 f\cdot\nu}{\sigma^2}
  \quad\text{on }\partial\Omega.
  \label{eq:conormal}
\end{equation}
For reflected VE dynamics with $f=0$, this reduces to the   Neumann condition. For reflected VP dynamics, however, $f\cdot\nu$ is generally nonzero, so imposing $\partial_\nu\log p=0$ would be inconsistent with the forward no-flux PDE. Boundary-aware score parameterizations should  enforce or approximate the co-normal relation \eqref{eq:conormal}, or  penalize the normal component of the \emph{learned velocity}
\begin{equation}
  v_\theta\cdot\nu
  =\left(f-\tfrac12\sigma^2s_\theta\right)\cdot\nu,
\end{equation}
rather than zeroing the score normal component independently of the drift.
This has a useful computational consequence: in the regular regime, one does not need to solve a reflected ODE or add projection/reflection events to the PF-ODE. Confinement follows from tangency of the velocity and the regular Lagrangian-flow structure. This separates the numerical treatment of the reverse SDE, which retains reflection, from the deterministic PF-ODE sampler, which does not.
\vspace{-10pt}
\subsection{Endpoint and boundary failure modes}
Two failure mechanisms clarify what a constrained sampler must control.
\paragraph{Data-endpoint singularity.}
Reflected Brownian motion started from a compactly supported density inside an interval has a unique smooth positive density for every $t>0$ and satisfies tangency at the walls, yet the score in the initial vacuum still behaves like $1/t$ \citep{cont2026reflected}. The PF-ODE has no regular Lagrangian flow from time zero, while it is well posed on every $[\delta,T]$ \citep{cont2026reflected}. Thus reflection does not cure the data-endpoint pathology; early stopping remains necessary for compactly supported data.
\vspace{-10pt}
\paragraph{Boundary-current compression.}
For more general no-flux density--flux pairs, if the density degenerates at the boundary then $p\,v\cdot\nu=0$ no longer implies $v\cdot\nu=0$. A boundary current can coexist with exact marginal transport and confined integral curves while producing unbounded compression and no regular Lagrangian flow \citep{cont2026reflected}. Uniformly elliptic reflected VE/VP models after positive-time regularization avoid this mechanism because their densities are positive up to the boundary, but the example is a warning for learned   constrained flows: preserving mass and keeping sample paths inside the domain are not sufficient for the stability of the deterministic sampler.

Geometry also enters through score regularity. Smooth convex domains are favorable for Neumann/co-normal regularity, whereas nonconvex or nonsmooth boundaries can degrade second-derivative estimates near the wall \citep{cont2026reflected}. Domain choice and boundary representation are therefore part of the conditioning of a reflected score model, not merely semantic constraints on the output space.\vspace{-10pt}
\section{Implications: Design principles for diffusion samplers}
\label{sec:design}
These results suggest treating the PF-ODE as a learned dynamical system, not a   by-product of score estimation. The central object is the learned probability-flow velocity, not the score in isolation.\vspace{-10pt}
\subsection{Regularize the quantity that is actually integrated}
The training objective should distinguish   score accuracy from regularity of the  velocity. One can use
\begin{equation}
\mathcal L_{\mathrm{PF}}
=\mathcal L_{\mathrm{DSM}}
+\lambda_J\mathcal R_J(v_\theta)
+\lambda_{\mathrm{div}}\mathcal R_{\mathrm{div}}(v_\theta)
+\lambda_g\mathcal R_{\mathrm{growth}}(v_\theta)
+\lambda_{\partial}\mathcal R_{\partial}(v_\theta),
\label{eq:designloss}
\end{equation}
where the last term is used only for constrained models. The role of each term follows directly from the flow conditions. $\mathcal R_J$ controls a Sobolev/Jacobian norm, $\mathcal R_{\mathrm{div}}$ controls the divergence component that drives compression, and $\mathcal R_{\mathrm{growth}}$ prevents rapid escape to spatial regions in which neither the score nor its error is controlled. For generation in one time direction, the divergence penalty may be one-sided; for invertible generation/encoding it should control both signs. Spectral normalization and Jacobian-trace penalties are concrete mechanisms for implementing this \citep{cont2026mimicking}.
For reflected models, the natural boundary regularizer is not $\|s_\theta\cdot\nu\|^2$ in general, but
\begin{equation}
  \mathcal R_{\partial}(v_\theta)
  =\E_{(x,t)\sim q_{\partial}}\!
  \left|\left(f(x,t)-\tfrac12\sigma(t)^2s_\theta(x,t)\right)\!\cdot\nu(x)\right|^2,
  \label{eq:boundaryreg}
\end{equation}
for a suitable distribution $q_{\partial}$ on boundary/time pairs. This enforces the  co-normal condition and reduces to a zero-normal-score condition only when $f\cdot\nu=0$. It is particularly relevant when a score network is trained in ambient coordinates and has no built-in knowledge of the domain geometry.

\begin{figure}[!htbp]
  \centering
  \begin{minipage}[t]{0.49\linewidth}
    \centering
    \includegraphics[width=\linewidth]{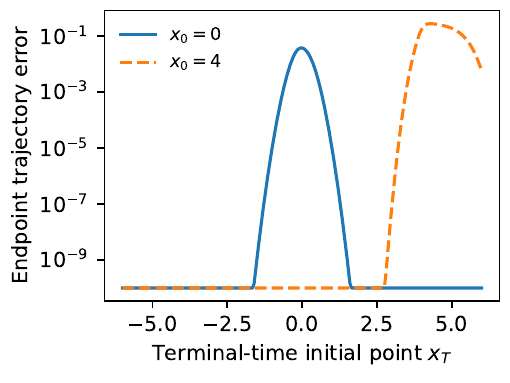}
    \par\small (a) Matched-DSM trajectory errors
  \end{minipage}\hfill
  \begin{minipage}[t]{0.49\linewidth}
    \centering
    \includegraphics[width=\linewidth]{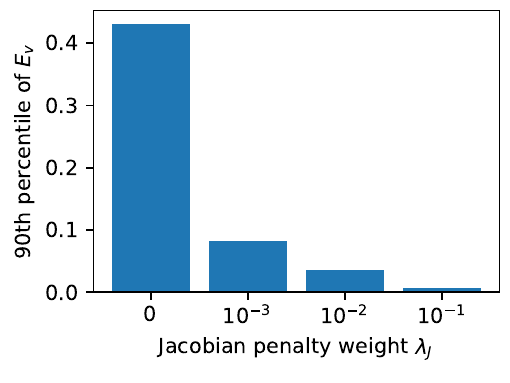}
    \par\small (b) Tail-mode regularization
  \end{minipage}
  \caption{Low-density diagnostics and a controlled regularization test.
  (a) Endpoint trajectory errors for the   score perturbations with
  $E_\theta^2=10^{-3}$; 
  (b)   $90$th percentile of unweighted velocity error over $5000$
  noisy-score regressions decreases with Jacobian penalty (matched DSM error not imposed).}
  \label{fig:score-error-numerics}
\end{figure}

\paragraph{Numerical illustration} To test whether such regularization can suppress an instability that is weakly identified by score matching, we consider the one-dimensional VE model \(p_t=N(0,1+t)\) and a learned score \(s_a=s+a h\), where \(h(x)=\exp(-(x-4)^2/(2w^2))\) is localized in a low-density region. With \(N=1024\) noisy score observations, the coefficient \(a\) is poorly identified by the density-weighted loss. Over 5000 repetitions, adding a spatially uniform Jacobian penalty reduces the 90th percentile of the unweighted velocity error \(E_v\) from \(0.431\) at \(\lambda_J=0\) to \(0.0825,0.0351,\) and \(0.00678\) for \(\lambda_J=10^{-3},10^{-2},10^{-1}\), respectively (see Figure \ref{fig:score-error-numerics}b). This   example illustrates that 
velocity regularization can suppress modes that receive little weight under the training distribution and substantially reduce the stability-relevant velocity error. 
 Details  are given in Appendix \ref{sec:numerics}.
\vspace{-20pt}
\subsection{Use density-stratified validation}\vspace{-5pt}
The inverse-density factor in \eqref{eq:mismatch} suggests that    global denoising-score-matching   is a poor diagnostic for ODE sampling. A more informative report stratifies validation points into bins according to   $\log p_t(x)$ or another proxy for density and evaluates   score   and velocity/Jacobian error in each bin. The same principle applies at the boundary of a constrained domain: evaluate tangency and compression as a function of distance to $\partial\Omega$. Such diagnostics target  regions that density weighting tends to hide.
\paragraph{Numerical illustration} In the   Gaussian example above, we perturb the  score by a localized bump of width \(w=0.2\), centered either at \(x_0=0\) or at the low-density location \(x_0=4\), and choose the amplitudes so that both perturbations have the same integrated DSM error \(E_\theta^2=10^{-3}\). Despite identical DSM loss, the tail perturbation produces a \(12.3\times\) larger unweighted velocity error, a \(152\times\) larger excess Jacobian penalty, and a \(15.9\times\) larger mean endpoint trajectory error on a uniform validation grid (Figure \ref{fig:score-error-numerics}a). Thus a global density-weighted score metric can make two models appear equivalent while density-stratified or trajectory-level diagnostics distinguish their PF-ODE behavior.

Trajectory-level tests are also important. If $Z_\theta$ is used for deterministic decoding or likelihood evaluation, one should test forward--reverse round trips, sensitivity to score perturbations, and  volume distortion along trajectories. These   are aligned with  properties controlled by regular Lagrangian-flow theory and can distinguish a low-loss score estimator from a stable ODE vector field.\vspace{-10pt}
\subsection{Choose schedules together with the endpoint cutoff}
For compactly supported data, early-stopping estimates depend on accumulated noise variance $\varsigma_t^2$. In the VE case $\varsigma_t^2\asymp t$, representative score-Hessian bounds scale like $\varsigma_t^{-2}+R^2\varsigma_t^{-4}$, so integrated conditioning deteriorates rapidly as the sampler is extended toward $t=0$. The relevant design variable is therefore the pair \emph{(noise schedule, endpoint cutoff)} rather than either one alone. A schedule that injects enough noise to regularize the score quickly can make a given positive cutoff better conditioned, while an aggressive reduction of $\delta$ can erase this benefit. This perspective is complementary to recent schedule-design work based on Lipschitz conditioning \citep{chen2025schedule}.\vspace{-10pt}
\paragraph{Numerical illustration} (Appendix \ref{sec:numerics}) The cutoff tradeoff is explicit for the VE model \(X_0=(Y,0)\), \(Y\sim N(0,1)\), in \(\mathbb R^{1+c}\). In the \(c\) normal directions the reverse PF-ODE from \(T=1\) to \(\delta\) is \(z_\delta=\sqrt{\delta}\,z_T\). Hence the residual normal scale decreases as \(\sqrt{\delta}\), while inverse sensitivity grows as \(\delta^{-1/2}\) and normal volume contracts by \(\delta^{c/2}\). For example, at \(\delta=10^{-2}\) the residual normal scale is \(0.1\), but   inverse sensitivity is already \(10\); for codimension \(c=8\), the normal volume factor is \(10^{-8}\). Thus improved endpoint fidelity is accompanied by   deteriorating invertibility even at positive cutoff.

Here the normal contribution to the Wasserstein-2 distance from the singular target is exactly $\sqrt{c\delta}$; at $\delta=10^{-2}$ this equals 0.283 for c=8, while the inverse normal sensitivity is 10. Thus finite-cutoff distributional error can be small while the ambient deterministic map is already strongly compressed.
This is arguably one of the central messages: distributional approximation can improve toward the manifold while ambient deterministic invertibility deteriorates.

For reflected models the same principle interacts with geometry. Parabolic regularization gives a smooth positive density up to a smooth boundary, but   constants depend on  boundary regularity and can deteriorate at corners or nonconvex singularities. If the constraint set is a modeling choice,  a smooth geometric representation can therefore improve   score regularity and deterministic sampling.
\begin{table}[t]
\caption{Theory-to-design translation for deterministic diffusion samplers.}
\label{tab:design}
\centering
\small
\begin{tabularx}{\textwidth}{>{\raggedright\arraybackslash}p{0.2\textwidth} >{\hsize=0.75\hsize}X >{\hsize=1.25\hsize}X} 
\toprule
Requirement/ failure mode & Analytic diagnostic & Design consequence \\
\midrule
PF-ODE existence & Sobolev/BV regularity, growth, $[\nabla\!\cdot v_\theta]^-$ control & Regularize Jacobian/divergence and network growth; do not rely on score loss alone. \\
Reverse/invertible flow & Both $[\nabla\!\cdot v_\theta]^-$ and $[\nabla\!\cdot v_\theta]^+$ & Stronger regularization for deterministic decoding, round trips, and likelihood evaluation. \\
Under the manifold hypothesis & Normal score/compression constants diverge as $t\downarrow0$ & Train/sample on $[\delta,T]$ and use a terminal denoising or limiting step; tune $\delta$ jointly with  schedule. \\
Low-density regions & $E_v$ depends on an unweighted norm; conversion from score loss uses $\int 1/p_t$ & Add localization or low-density diagnostics/regularization; validate trajectories, not only DSM loss. \\
Reflected boundary & $v_\theta\cdot\nu\approx0$; generally $s_\theta\cdot\nu\neq0$ & Enforce   co-normal relation for the   drift--score field; avoid generic zero-normal-score constraints. \\
Boundary degeneracy & Loss of positivity can permit large compression near $\partial\Omega$ & Maintain positive-time regularization; monitor boundary density, normal trace, and compression. \\
\bottomrule
\end{tabularx}
\end{table} 
These design principles, summarized in  Table~\ref{tab:design}, suggest that validation protocols for ODE-based diffusion models should, in addition to score loss and sample-quality metrics, evaluate: (i) the time profile of Jacobian/divergence norms of $v_\theta$; (ii) empirical volume compression along trajectories; (iii) sensitivity of the sampler to decreasing $\delta$; (iv) score/velocity errors stratified by density level; and, for constrained models, (v) the boundary residual $|v_\theta\cdot\nu|$. These diagnostics target   quantities that affect the  flow   and can reveal failures invisible to density-weighted training loss.

The results also separate two modeling choices that are often conflated. The reverse SDE is a stochastic process whose score error is naturally averaged under its evolving law, while the PF-ODE is a deterministic vector field whose global trajectories can be affected by low-density regions. A score architecture and schedule optimized for SDE sampling need not be optimal for ODE sampling. If deterministic decoding, fast ODE generation, or exact likelihoods are central goals, PF-ODE conditioning should be included explicitly in training and schedule selection.

\appendix
\section{Some mathematical results on ODE samplers for diffusion models}
\label{app:formal}
This appendix summarizes results from \cite{cont2026mimicking,cont2026reflected}.
For mathematical definitions we refer to \citet{Ambrosio2008}.
\subsection{PF-ODE in the whole space}
Consider \eqref{eq:sde} with scalar uniformly nondegenerate $\sigma(t)\geq \sigma_{\min}>0$, and let $p_t$ be its marginal density. Suppose the drift is locally bounded variation (BV) in space with integrable growth, its distributional divergence is absolutely continuous with an integrable one-sided $L^\infty$ bound, and the score $\nabla\log p_t$ has corresponding Sobolev/BV regularity, linear growth, and an integrable upper bound on $\Delta\log p_t$. Then the velocity \eqref{eq:velocity} generates a unique regular Lagrangian flow and transports $p_0$ to $p_t$. If the opposite sign of $\nabla\cdot v$ is also integrably bounded in $L^\infty$, the time-reversed regular Lagrangian flow exists and is the a.e. inverse of the forward flow.

For linear VP/VE diffusions with compactly supported initial law, no density or score at time zero is required: for each $\delta>0$, Gaussian regularization yields the required score bounds on $[\delta,T]$. In the VE scaling $\varsigma_t^2\asymp t$, worst-case bounds grow polynomially in $\delta^{-1}$, which explains the deterioration toward the data endpoint.

\subsection{Learned-score stability}

Let $Z^\delta$ and $Z_\theta^\delta$ be the exact and learned regular Lagrangian flows, both initialized as the identity at time $\delta$. Under local $W^{1,p}$ regularity ($p>1$), growth, and compression bounds, for $x$ in a ball $B_r$ one obtains an estimate of the form
\begin{equation}
 \cL^d\!\left(\left\{x\in B_r:\left|Z^\delta(T,x)-Z_\theta^\delta(T,x)\right|>\eta\right\}\right)
 \le \frac{C_p+L_\theta E_v/\lambda}{\log(1+\eta/\lambda)},
\end{equation}
where $E_v$ is defined in \eqref{eq:Ev}, $L_\theta$ is the compression constant of the learned flow, and $C_p$ depends on the Sobolev norm of the exact velocity on a confinement ball. Taking $\lambda=\sqrt{E_v}$ yields convergence in measure as $E_v\to0$ with fixed regularity/compression constants. Equation \eqref{eq:mismatch} relates $E_v$ to the denoising score-matching loss.

\subsection{PF-ODE in a bounded domain}
Let $\Omega\subset \mathbb{R}^d$ be a bounded smooth domain and consider the normally reflected diffusion \eqref{eq:reflectedsde}. Its probability flux is $J=fp-\frac12\sigma^2\nabla p$. If, on $[\delta,T]$, $p$ is positive up to the boundary and the velocity $v=J/p$ has the interior/collar BV regularity \citep{crippa2008} and one-sided divergence bounds required for a regular Lagrangian flow \citep{Ambrosio2008}, then the no-flux condition gives a vanishing normal trace of $v$. Extending $v$ by zero outside $\Omega$ introduces no singular boundary divergence term, and the resulting RLF stays in $\overline\Omega$ and transports $p_\delta$ to $p_t$.

For $C^{2,\alpha}$ domains and sufficiently regular drift, parabolic boundary regularity verifies these conditions for every $\delta>0$ and arbitrary initial law. The reflected VE and VP diffusions used in constrained score-based modeling satisfy the resulting early-stopped transport statement. The constants degenerate as $\delta\downarrow0$.

\section{Practical checklist for a PF-ODE implementation}

For an implementation intended to use deterministic PF-ODE sampling, the theory motivates the following checks in addition to ordinary score-model evaluation:
\begin{enumerate}[leftmargin=*]
  \item report the data-endpoint truncation $\delta$ and the terminal denoising procedure;
  \item measure Jacobian/divergence norms of $v_\theta$ over time and across density quantiles;
  \item test sensitivity of sample trajectories and likelihood estimates to the ODE solver and to perturbations of $s_\theta$;
  \item if invertibility is used, test both forward and reverse compression rather than forward generation alone;
  \item for reflected models, evaluate $v_\theta\cdot\nu$ on boundary samples and use the drift-dependent co-normal target;
  \item stratify score error by density level so that tail or boundary errors are not hidden by density weighting.
\end{enumerate}

\subsection{Numerical experiments}
\label{sec:numerics}
We report three numerical experiments that isolate the mechanisms behind the
design principles. Exact reference scores allow us to distinguish score approximation,
flow conditioning, and numerical integration error. 
\paragraph{Jacobian regularization of a poorly identified tail mode.}
We   test a simplified instance of the regularization principle in
\eqref{eq:designloss}. Retain $h=h_4$ and the model $s_a=s+a h$ above, whose true
coefficient is $a=0$. For each of $5000$
independent training sets, draw $N=1024$ pairs with $t_i$ uniform on $[0.05,1]$ and
$x_i\sim p_{t_i}$, and use targets $s(x_i,t_i)+\varepsilon_i$, where
$\varepsilon_i\sim\mathcal N(0,0.2^2)$ independently.   Penalizing the excess velocity Jacobian under
a spatially uniform reference distribution gives
\begin{equation}
  \mathcal R_{J,\mathrm{err}}(a)=K_Ja^2,
  \qquad
  K_J=\frac{1}{2R}\int_{-R}^{R}\left|\tfrac12 h'(x)\right|^2\,\dd x,
  \qquad R=6.
  \label{eq:numerical-tail-penalty}
\end{equation}
Minimizing the empirical squared regression loss plus
$\lambda_J\mathcal R_{J,\mathrm{err}}(a)$ gives
\begin{equation}
  \widehat a_{\lambda_J}
   =\frac{N^{-1}\sum_{i=1}^{N}h(x_i)\varepsilon_i}
          {N^{-1}\sum_{i=1}^{N}h(x_i)^2+\lambda_JK_J}.
  \label{eq:numerical-tail-estimator}
\end{equation}
The same training sets are used for all penalty weights. Since the tail basis is rarely
sampled, its coefficient is poorly identified without regularization.
Table~\ref{tab:jacobian-numerics} and Figure~\ref{fig:score-error-numerics}(b) show that the
penalty suppresses this residual mode: the $90$th percentile of $E_v$ decreases from
$0.4308$ without regularization to $0.03510$ at $\lambda_J=10^{-2}$ and $0.006779$ at
$\lambda_J=10^{-1}$. Population DSM error also decreases.
\begin{table}[!htbp]
  \caption{Tail-mode regularization across $5000$ independent training sets. Each column
  reports a separate empirical $90$th percentile. The penalty is spatially normalized as
  in \eqref{eq:numerical-tail-penalty}, whereas $E_v$ and $E_\theta^2$ retain the integrated
  definitions in \eqref{eq:Ev} and \eqref{eq:scoreloss}.}
  \label{tab:jacobian-numerics}
  \centering
  \small
  \begin{tabular}{rrrr}
    \toprule
    $\lambda_J$ & $|\widehat a_{\lambda_J}|$ & $E_\theta^2$ & $E_v$ \\
    \midrule
    $0$        & $1.8091$  & $2.361\times10^{-3}$ & $0.4308$ \\
    $10^{-3}$  & $0.34648$ & $8.659\times10^{-5}$ & $0.08251$ \\
    $10^{-2}$  & $0.14741$ & $1.567\times10^{-5}$ & $0.03510$ \\
    $10^{-1}$  & $0.02847$ & $5.846\times10^{-7}$ & $0.006779$ \\
    \bottomrule
  \end{tabular}
\end{table}

\paragraph{Equal DSM error, different velocity and trajectory errors.}
To isolate \eqref{eq:mismatch}, consider $p_t=\mathcal N(0,1+t)$ on
$[\delta,T]=[0.05,1]$, with $s(x,t)=-x/(1+t)$ and $v=-s/2$. Define
\begin{equation}
  s_\theta(x,t)=s(x,t)+A h_{x_0}(x),
  \qquad
  h_{x_0}(x)=\exp\!\left(-\frac{(x-x_0)^2}{2w^2}\right),
  \qquad w=0.2.
  \label{eq:numerical-score-bump}
\end{equation}
We compare $x_0=0$ and $x_0=4$, choosing $A$ by quadrature so that both perturbations have
$E_\theta^2=10^{-3}$ in \eqref{eq:scoreloss}. Here ``DSM error'' denotes that population
score error, not a noisy empirical denoising loss. Besides $E_v$ on $B_R=[-6,6]$, we evaluate
the excess Jacobian error
$J_{\rm err}:=\int_\delta^T\int_{-6}^{6}|\partial_x(v_\theta-v)|^2\,\dd x\,\dd t$.
Reverse trajectories are initialized at $241$ equally spaced points $x_T\in[-6,6]$ and
compared with the exact endpoint $x_T\sqrt{(1+\delta)/(1+T)}$.

\begin{table}[!htbp]
  \caption{Matched population DSM error for central and tail score perturbations.
  Trajectory errors are evaluated at $t=\delta$; the mean is an arithmetic mean over the
  uniform spatial grid, not an expectation under $p_T$. The Jacobian error is the
  unnormalized space--time integral $J_{\rm err}$.}
  \label{tab:dsm-velocity-numerics}
  \centering
  \small
  \begin{tabular}{lrr}
    \toprule
    Diagnostic & Central defect ($x_0=0$) & Tail defect ($x_0=4$) \\
    \midrule
    Score amplitude $A$ & $0.09560$ & $1.17745$ \\
    Population DSM error $E_\theta^2$ & $0.001$ & $0.001$ \\
    Unweighted velocity error $E_v$ & $0.02276$ & $0.28039$ \\
    Excess Jacobian error $J_{\rm err}$ & $0.009618$ & $1.45902$ \\
    Mean grid trajectory error & $0.001854$ & $0.02953$ \\
    Maximum grid trajectory error & $0.03788$ & $0.28037$ \\
    \bottomrule
  \end{tabular}
\end{table}

Despite equal DSM error, the tail perturbation has a $12.3$-fold larger $E_v$, a $151.7$-fold
larger $J_{\rm err}$, and a $15.9$-fold larger mean grid trajectory error
(Table~\ref{tab:dsm-velocity-numerics}). Figure~\ref{fig:score-error-numerics}(a) shows the
spatial localization of the discrepancies. This supports spatially resolved and low-density
validation of the deterministic field.

\begin{figure}[!htbp]
  \centering
  \begin{minipage}[t]{0.49\linewidth}
    \centering
    \includegraphics[width=\linewidth]{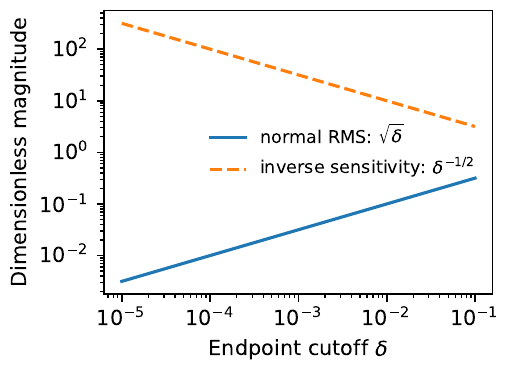}
    \par\small (a) Fidelity and inverse sensitivity
  \end{minipage}\hfill
  \begin{minipage}[t]{0.49\linewidth}
    \centering
    \includegraphics[width=\linewidth]{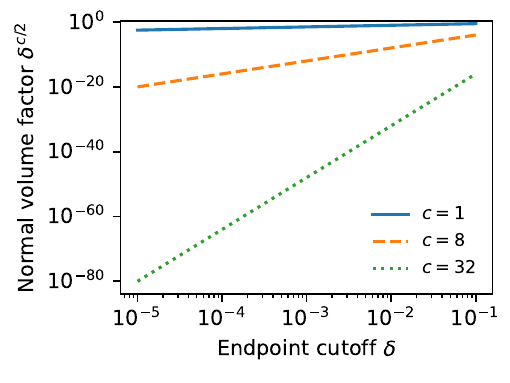}
    \par\small (b) Normal volume collapse
  \end{minipage}
  \caption{Exact-score endpoint diagnostics for the flat-manifold VE model, with $T=1$ and
  $\delta\in[10^{-5},10^{-1}]$. Decreasing the normal residual improves fidelity to the data
  subspace while increasing the sensitivity of the inverse map. Normal volume contraction
  is amplified by codimension $c$; these curves follow the exact expressions in
  \eqref{eq:numerical-flat-diagnostics}.}
  \label{fig:endpoint-numerics}
\end{figure}

\paragraph{Endpoint fidelity versus conditioning.}
As a closed-form analogue of Proposition~\ref{prop:sphere}, let
$X_0=(Y,0_c)\in\R^{1+c}$, with $Y\sim\mathcal N(0,1)$ and $c\geq1$, and set
$X_t=X_0+W_t$. Then
$p_t=\mathcal N(0,1+t)\otimes\mathcal N(0,tI_c)$ and
$v(y,z,t)=(y/[2(1+t)],z/[2t])$. The reverse PF-ODE map from $T$ to $\delta$ is
\begin{equation}
  \Phi_{\delta,T}(y,z)
  =\left(\sqrt{\frac{1+\delta}{1+T}}\,y,
          \sqrt{\frac{\delta}{T}}\,z\right).
  \label{eq:numerical-flat-flow}
\end{equation}
For initialization from $p_T$, the normal component $\Phi^z_{\delta,T}$ satisfies
\begin{equation}
  \left(\frac{\E|z_\delta|^2}{c}\right)^{1/2}=\sqrt{\delta},
  \qquad
  \left\|\bigl(D_z\Phi^z_{\delta,T}\bigr)^{-1}\right\|
    =\sqrt{\frac{T}{\delta}},
  \qquad
  \det D_z\Phi^z_{\delta,T}
    =\left(\frac{\delta}{T}\right)^{c/2}.
  \label{eq:numerical-flat-diagnostics}
\end{equation}
Figure~\ref{fig:endpoint-numerics} shows the resulting fidelity--conditioning tradeoff for
$T=1$. At $\delta=10^{-2}$, the normal RMS per coordinate is $0.1$ and the inverse normal
sensitivity is $10$, while the normal volume factors are $10^{-8}$ for $c=8$ and $10^{-32}$
for $c=32$.  
For the scalar equation $\dot z=z/(2t)$ with $z(1)=1$, adaptive Runge-Kutta45 with relative tolerance
$10^{-6}$ uses $98$, $194$, $290$, and $470$ function
evaluations at cutoffs $10^{-1}$, $10^{-2}$, $10^{-3}$, and $10^{-5}$, respectively.
Across the $41$ tested cutoffs, the relative endpoint error remains below $2.1\times10^{-6}$.
This example exhibits increasing solver work and inverse sensitivity: the exact reverse map is contractive in the normal directions.

\paragraph{Implementation details}
DSM calibration uses trapezoidal quadrature with
$501$ time nodes and $16001$ spatial nodes on $[-8,8]$; $E_v$ and $J_{\rm err}$ use $30001$
spatial nodes on $[-6,6]$, and $K_J$ uses $100001$ nodes. The matched-DSM trajectories use
RK45 with relative tolerance $10^{-8}$, absolute tolerance $10^{-10}$, and maximum step
$0.01$. The regression experiment uses NumPy's default random generator with seed $42$.
  The regularization test
uses a known reference score and a single coefficient.  

\end{document}